\documentclass{article}
\pdfoutput=1
\usepackage[utf8]{inputenc}
\usepackage{authblk}
\usepackage{setspace}
\usepackage[margin=1.25in]{geometry}
\usepackage{graphicx}
\usepackage{subcaption}
\usepackage{amsmath}
\usepackage{bm}
\usepackage{amssymb}
\usepackage{tabularx}
\usepackage{booktabs}
\usepackage{multirow}
\usepackage[table]{xcolor}

\definecolor{mygre}{RGB}{0,128,0}  
\definecolor{myred}{RGB}{200,0,0}    
\definecolor{mygra}{RGB}{80,80,80}    

\usepackage[style=nejm, 
citestyle=numeric-comp,
sorting=none]{biblatex}
\title{KidneyTalk-open: No-code Deployment of a Private Large Language Model with Medical Documentation-Enhanced Knowledge Database for Kidney Disease}

\author[1,2]{Yongchao Long}
\author[4,7]{Chao Yang}
\author[2]{Gongzheng Tang}
\author[4]{Jinwei Wang}
\author[9]{Zhun Sui}
\author[1,3,*]{Yuxi Zhou}
\author[2,8,*]{Shenda Hong}
\author[2,4,5,6,7,*]{Luxia Zhang}

\affil[1]{Department of Computer Science, Tianjin University of Technology, Tianjin, China.}
\affil[2]{National Institute of Health Data Science, Peking University, Beijing, China}
\affil[3]{Institute of Internet Industry, Tsinghua University, Beijing, China}
\affil[4]{Renal Division, Department of Medicine, Peking University First Hospital, Beijing, China}
\affil[5]{Research Units of Diagnosis and Treatment of Immune-Mediated Kidney Diseases, Chinese Academy of Medical Sciences, Beijing, China}
\affil[6]{State Key Laboratory of Vascular Homeostasis and Remodeling, Peking University, Beijing, China}
\affil[7]{Center for Digital Health and Artificial Intelligence, Peking University First Hospital, Beijing, China}
\affil[8]{Department of Emergency Medicine, Peking University First Hospital, Beijing, China}
\affil[9]{Renal Department, Peking University People's Hospital, Beijing, China}

\affil[*]{Address correspondence to: zhanglx@bjmu.edu.cn; joy\_yuxi@pku.edu.cn; hongshenda@pku.edu.cn}

\date{}

\begin{document}

\maketitle


\begin{abstract} 
\textbf{Background:} Privacy-preserving medical decision support for kidney disease requires localized deployment of large language models (LLMs) while maintaining clinical reasoning capabilities. Current solutions face three challenges: 1) Cloud-based LLMs pose data security risks; 2) Local model deployment demands technical expertise; 3) General LLMs lack mechanisms to integrate medical knowledge. Retrieval-augmented systems also struggle with medical document processing and clinical usability. \textbf{Methods:} We developed KidneyTalk-open, a desktop system integrating three technical components: 1) No-code deployment of state-of-the-art (SOTA) open-source LLMs (such as DeepSeek-r1, Qwen2.5) via local inference engine; 2) Medical document processing pipeline combining context-aware chunking and intelligent filtering; 3) Adaptive Retrieval and Augmentation Pipeline (AddRep) employing agents collaboration for improving the recall rate of medical documents. A graphical interface was designed to enable clinicians to manage medical documents and conduct AI-powered consultations without technical expertise. \textbf{Results:} Experimental validation on 1,455 challenging nephrology exam questions demonstrates AddRep's effectiveness: achieving 29.1\% accuracy (+8.1\% over baseline) with intelligent knowledge integration, while maintaining robustness through 4.9\% rejection rate to suppress hallucinations. Comparative case studies with the mainstream products (AnythingLLM, Chatbox, GPT4ALL) demonstrate KidneyTalk-open's superior performance in real clinical query. \textbf{Conclusions:} KidneyTalk-open represents the first no-code medical LLM system enabling secure documentation-enhanced medical Q\&A on desktop. Its designs establishes a new framework for privacy-sensitive clinical AI applications. The system significantly lowers technical barriers while improving evidence traceability, enabling more medical staff or patients to use SOTA open-source LLMs conveniently.
\end{abstract}

\section{Introduction}

Research on large language models (LLMs) in the medical domain has achieved significant progress in recent years. Representative models such as Med-PaLM 2 (340B parameters) \cite{singhal2025toward} and MedFound (176B parameters) \cite{liu2025generalist} demonstrate professional advantages in medical Q\&A and clinical text processing tasks. These models exhibit excellent performance on standardized medical benchmarks through domain-adaptive fine-tuning of medical corpora. However, their clinical application faces dual dilemmas: On one hand, ultra-large parameter models are difficult to deploy in resource-constrained environments like personal computers; On the other hand, domain specialization leads to significant degradation in general language capabilities \cite{dou2024loramoe}, limiting their application scope in diverse medical scenarios. Recent studies reveal that general LLMs inherently possess strong medical reasoning capabilities \cite{jeong2024limited}, suggesting possibilities for building lightweight medical intelligence systems.

Although general LLMs\cite{dubey2024llama,yang2024qwen2,liu2024deepseek,guo2025deepseek} demonstrate excellent cross-domain reasoning capabilities, their medical applications still face three key limitations: First, most commercial models only provide cloud API services, failing to meet the strict patient privacy protection requirements of medical institutions; Second, local deployment of open-source LLMs requires complex command-line operations and parameter configuration, creating significant technical barriers; Third, the timeliness of model training data makes it difficult to integrate the latest clinical guidelines or medical discoveries. Research shows that Retrieval-Augmented Generation (RAG) technology \cite{fan2024survey,luo2024development,kresevic2024optimization,singhal2025toward} can effectively mitigate model hallucination and enhance knowledge timeliness through dynamic association with external knowledge databases\cite{trivedi2022interleaving,ram2023context,asai2023retrieval,xiong2024benchmarking}. However, existing open-source systems have not yet well reached clinical utility standards in medical document processing, privacy protection, and interaction design.

To address these challenges, this paper introduces KidneyTalk-open - the first no-code deployable private medical LLM system, with innovations in three dimensions: 1) Highly integrated deployment of advanced open-source models with medical knowledge databases in local environments, achieving complete data lifecycle management from document parsing to reasoning output; 2) A pipeline for multi-agent collaboration based on LLM is designed to enhance the utilization rate of knowledge; 3) Development of a zero-technical-barrier graphical interface enabling clinical practitioners to manage medical knowledge databases and obtain decision support without programming skills. 

The validation framework of KidneyTalk-open is systematically designed across two dimensions to demonstrate its clinical superiority. First, through rigorous evaluation on the Chinese Nephrology Medical Exam MCQ dataset (CNME-MCQ), our proposed multi-agent collaboration based on LLM pipeline achieves 29.1\% overall accuracy, outperforming baseline methods by 8.1\% across four difficulty levels, while maintaining robust hallucination suppression with a 4.9\% rejection rate. Second, in comparative analysis with mainstream localized LLM applications (AnythingLLM, Chatbox, GPT4ALL) through a targeted diabetic kidney disease case study, KidneyTalk-open demonstrates comprehensive coverage of clinical guidelines by addressing 5/6 essential management dimensions - including unique achievement of stage-specific treatment personalization through eGFR-adaptive protocol retrieval - whereas competitors showed critical deficits, with at most 2 dimensions adequately addressed and fundamental failures in guideline integration.

KidneyTalk-open can be downloaded for free from our website\footnote{\url{https://github.com/PKUDigitalHealth/KidneyTalk-open}}, and the development code is open source. The paper structure is organized as follows: Section 2 details the core technical solutions; Section 3 demonstrates workflows through clinical cases; Section 4 presents systematic validation of technical components and comparative analysis with mainstream tools; Followed by discussion of clinical value and future directions.

\section{Design and Development}

KidneyTalk-open is a fully privatized large language model system tailored for medical Q\&A, integrating local knowledge databases and model inference engines. Its architecture prioritizes clinical workflow security and usability requirements, enabling desktop installation with low hardware demands. Through deep customization of existing mature technologies, the system achieves organic integration of three core functionalities: intelligent medical document management, instant medical knowledge retrieval, and evidence-based decision support. The system's technical breakthroughs manifest in three dimensions:

First, by integrating high-performance LLM inference engines, it achieves automatic model parameter configuration and hardware adaptation, supporting one-click model acquisition through graphical interfaces (default configurations include DeepSeek-r1 and Qwen2.5, completely eliminating command-line operations and manual parameter tuning.

Second, developing a medical document-based knowledge database module that splits various format medical documents into semantic units, automatically filters redundant content through customizable agents, and constructs an efficient HNSW vector-indexed retrieval system enabling sub-second precise positioning of massive medical literature.

Third, creating a conversational clinical decision support interface that transforms complex processes like evidence-based guideline interpretation, differential diagnosis reasoning, and treatment plan validation into natural language interactions through pre-configured knowledge retrieval agents, allowing clinicians to obtain trustworthy decision support without technical expertise. The system strictly adheres to medical data lifecycle management standards, with all computations performed locally, providing dual guarantees for patient privacy and medical information security.

\subsection{Models Deployment and Inference}

KidneyTalk-open's foundation framework Jan \cite{janhq} is a mature open-source LLM client with a user-friendly interface. While Jan's default integrated Cortex \cite{cortex} inference engine encountered network connectivity issues for the users in China during model parameter downloads and had an incomplete model library, we replaced it with the more robust Ollama \cite{ollama} inference engine. Ollama provides stable network services crucial for downloading large LLM parameters and offers an advanced, comprehensive model library. Through Ollama's API integration, KidneyTalk-open achieves automatic model retrieval, eliminating complex command-line operations and model configurations - the key to its out-of-the-box usability.

During KidneyTalk-open initialization, three models are automatically downloaded from the Ollama repository. Download duration depends on network conditions.

\paragraph{Embedding Model}
BGE-M3 (1.2GB), based on the XLM-RoBERTa architecture, excels in dense, sparse, and multi-vector retrieval across 100+ languages, processing documents up to 8,192 tokens. As the core of our knowledge database module, it converts document content into vector representations for rapid semantic retrieval.

\paragraph{Text Generation Model}
Qwen2.5:7B (4.7GB) demonstrates superior performance across multiple domains, trained on 18 trillion tokens with supervised fine-tuning and RL optimization. We employ it as the base model for multiple agents.

\paragraph{Resoning Model}
DeepSeek-r1:7B (4.7GB), a distilled model fine-tuned from Qwen using DeepSeek-generated data, shows exceptional reasoning capabilities (achieving 55.5\% on AIME 2024), making it ideal for medical knowledge reasoning on resource-constrained hardware. Users with powerful desktop hardware can seamlessly upgrade to DeepSeek-r1:70B, which surpasses OpenAI's closed-source models. KidneyTalk-open's modular design allows effortless model replacement via Ollama without requiring technical expertise.

\begin{figure}[htbp]
    \centering
    \includegraphics[width=0.8\textwidth,page=1]{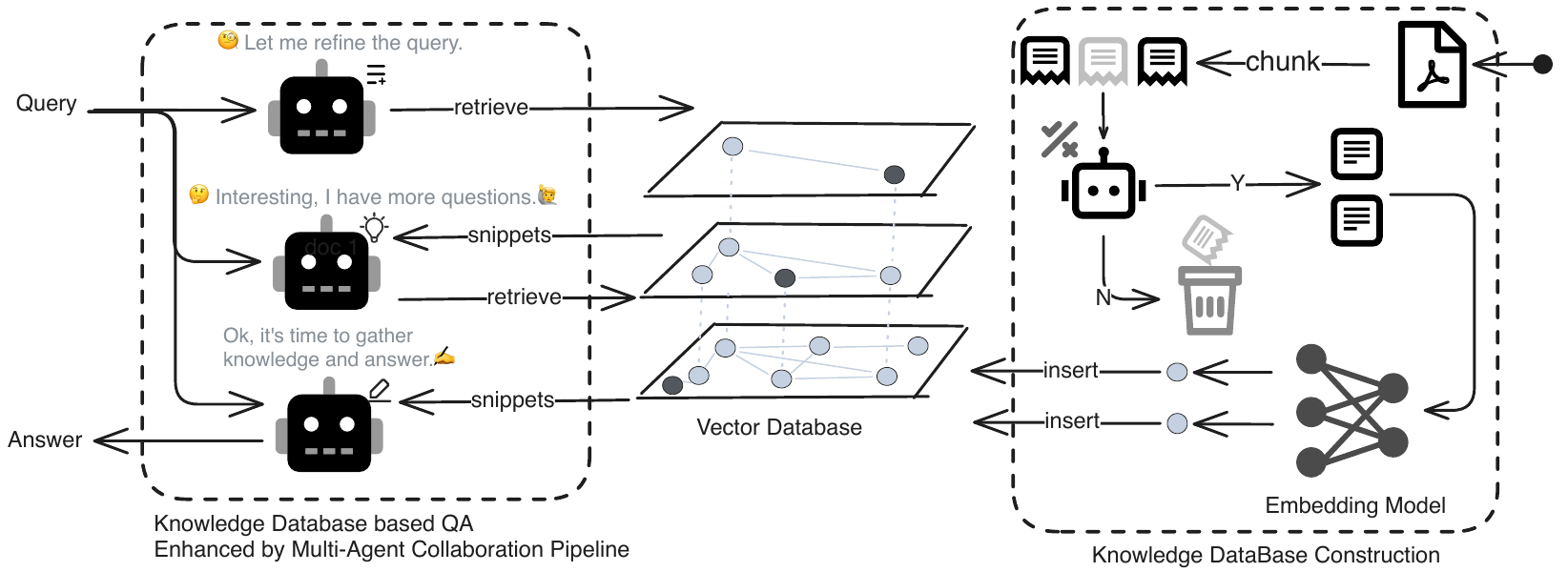}
    \caption{Schematic Diagram of the Core Design of KidneyTalk-open. \textbf{Right}: The knowledge database (KB) construction process. It commences with parsing PDF, Word, or Markdown documents into plain text format, proceeds to document chunking for generating knowledge snippets, then conducts knowledge snippet filtering, semantic embedding of the snippets, and concludes with storing the embedded vectors in a vector database. \textbf{Middle}: A schematic of the Hierarchical Navigable Small World (HNSW) vector database, a structure for efficient vector data storage and retrieval, which is employed to manage the knowledge vectors in the KB. \textbf{Left}: The Adaptive Retrieval and Augmentation Pipeline (AddRep) method proposed in this paper for enhancing medical document recall is introduced. It utilizes the collaboration of multiple agents (program modules with specific functions) to expand the user's question space. When a user submits a query, these agents perform operations such as query refinement and divergent thinking to comprehensively obtain relevant knowledge snippets, improving the accuracy and comprehensiveness of answers and ultimately providing an answer.}
    \label{fig:pipeline}
\end{figure}

\subsection{Knowledge Database Construction}

Constructing knowledge databases enables LLMs to incorporate external, structured medical knowledge, compensating for limitations in parametric knowledge. This enhancement improves answer accuracy and reliability by grounding responses in authoritative or up-to-date medical data \cite{singhal2025toward}. The underlying principle leverages the Transformer architecture's \cite{vaswani2017attention} autoregressive text generation through joint probability maximization:
\begin{equation}
p(x_{1}, \dots, x_{n}) = \prod_{i=1}^{n} p_{\theta}(x_{i} \mid x_{<i}),
\end{equation}
where $\theta$ denotes model parameters and $x_{<i}$ represents preceding tokens. Since context quality directly impacts LLM responses, we enhance medical Q\&A performance by integrating medical documents into the context window:
\begin{equation}
p(x_{1}, \dots, x_{n}) = \prod_{i=1}^{n} p(x_{i} \mid x_{<i}, R_C(x_{<i},\text{topk})).
\end{equation}
where $R_C(\cdot,\cdot)$ denotes the medical document retrieval function that recalls the $\text{topk}$ most relevant knowledge snippets. This addresses two critical challenges: 1) Limited context window sizes (typically 8,192 tokens) and KV cache constraints on desktop hardware; 2) The need for precise retrieval from extensive medical documents. Our knowledge database construction process, shown in Figure \ref{fig:pipeline} (right), includes document parsing, chunking, filtering, semantic embedding, and HNSW vector database integration.

\paragraph{1) Document Parsing \& Chunking}
KidneyTalk-open utilizes LangChain \cite{langchainjs} to parse PDF, Word, and TXT documents, applying automatic chunking with 512-token maximum length and 25\% overlap, approximating natural paragraph length.

\paragraph{2) Knowledge Snippets Filtering}
\label{sec:filter_agent}
We discovered that LangChain's automatic segmentation mechanism tends to generate knowledge snippets that are either blank or consist of too few words. Additionally, given the unstructured nature of PDF files, the parsing process results in knowledge snippets containing merely useless layout elements like headers and footers. Moreover, medical documents commonly incorporate references and other content lacking practical significance.

Consequently, we initially apply rules to filter out knowledge snippets that are blank or have fewer than 10 characters. Subsequently, we design an agent named \textit{Filter} for automatically differentiating irrelevant knowledge snippets. The default prompts of the \textit{Filter} classify references, headers, footers, and knowledge snippets devoid of medical knowledge as irrelevant ones and discard them. To expedite the discrimination task execution, the filter is configured to only return "Y" or "N". The function of knowledge snippet filtering lies in ensuring the content quality of the knowledge database, reducing noise, and thereby enhancing the recall rate. 

\paragraph{3) Semantic Embedding}
Immediately after that, knowledge snippets are transformed into vector representations through the text embedding model $\text{Embed}(\cdot)$:
\begin{equation}
\mathbf{v}_{ij} \in \mathbb{R}^{d} = \text{Embed}(t_{ij}^*) , 
\end{equation}
where $d$ is the dimension of the vector, and \(t_{ij}^*\) is the filtered knowledge snippet. The BGE-M3 can map indefinite-length knowledge snippets into 768-dimensional semantic vectors, so that during retrieval, the cosine similarity between the query vector and multiple knowledge snippet vectors in the knowledge database can be calculated in parallel, and the knowledge snippets related to the query can be calculated at the semantic level.

\paragraph{4) HNSW Vector Database}
The role of the vector database is to store vectors. We expect users to add millions of knowledge snippets to the local vector library, so it is necessary to ensure a high recall rate while ensuring that the retrieval speed is within the acceptable range of users. KidneyTalk-open uses HNSWLib\cite{hnswlib} as the vector library, a graph-structured vector database that combines retrieval efficiency and recall rate. The HNSW algorithm\cite{malkov2018efficient} achieves efficient approximate nearest neighbor search through a hierarchical small-world graph. Its design includes a hierarchical graph structure (the schematic diagram of HNSW is shown in the middle part of Figure \ref{fig:pipeline}), dynamic insertion, and heuristic retrieval strategies, ensuring excellent retrieval performance. In the hierarchical graph, the nodes in the upper layers are sparse, and the nodes in the bottom layers are dense. New nodes determine their levels through the probability distribution function:
\begin{equation}
\label{eq:hnsw}
P(l) = e^{-l \cdot \lambda}, \quad \lambda = \frac{1}{\log(M)}
\end{equation}
where $l$ denotes the level of the graph, $\lambda$ is the level probability factor, and the parameter $M$ represents the maximum number of neighbors that each node can have in the graph. It directly affects the density and navigability of the graph, and then affects the efficiency of index construction and the search process. Empirically, we obtain the best retrieval effect when setting the maximum level $l_{max}$ to 64 and $M$ to 8. The sparsity of the upper layers helps to quickly narrow the search scope, and the density of the bottom layers ensures the coverage ability. When inserting a new node, starting from the level entry point determined by the probability distribution function (Equation \ref{eq:hnsw}), the greedy and heuristic algorithms are used to find the optimal combination of neighbors until reaching the bottom layer. The heuristic algorithm ensures that the distance between neighbor nodes is greater than the distance from the target node to the new node, maintaining the sparsity of the graph, thereby improving the diversity of knowledge snippet retrieval. In the retrieval stage, the query point starts navigating from the highest-level entry point, narrows the search scope layer by layer, and finally conducts an extended search at the bottom layer, maintaining a set of candidate nodes until the stop condition is met. The hierarchical design and heuristic strategies of HNSW perform excellently in high-dimensional vector space and are applicable to the scenario of efficient document retrieval, which can be formally represented as:
\begin{equation}
\mathcal{T}_q = \{(t_i^q, s_i^q)\}_{i=1}^{\text{topk}} = R_C(q, \text{topk})
\end{equation}
where $\mathcal{T}_q$ represents the set of retrieved knowledge snippets, $t_i^q$ represents the knowledge snippet, $s_i^q$ represents the distance score from the query text $q$, and $\text{topk}$ is the expected number of knowledge snippets to be retrieved.

\subsection{Knowledge Q\&A}
\label{sec:addrep}

The performance of knowledge-based question answering systems is jointly determined by the recall rate of retrieval systems and the reasoning capabilities of LLMs. For the latter, KidneyTalk-open allows users to utilize state-of-the-art open-source LLMs, with our recommendation being the DeepSeek-r1:7B inference model. Section \ref{sec:val} demonstrates its distinction from the Qwen2.5:7B model. This section primarily focuses on enhancing KidneyTalk-open's retrieval recall rate based on the HNSW vector database.

The Adaptive Retrieval and Augmentation Pipeline (AddRep) method, which aims to enhance the response depth and diversity of LLMs in medical knowledge using agent technology, is inspired by RQ-RAG\cite{chan2024rq} and In-context RALM\cite{ram2023context}. It uses LLMs with commonsense understanding and reasoning capabilities to act as three different agents, who are in charge of query optimization, divergent thinking questioning, and knowledge integration for generating the final response. The AddRep workflow is illustrated in the left portion of Figure \ref{fig:pipeline}.

\paragraph{1) Query Refinement}
Upon receiving user query $q$, the \textit{Query Refinement Agent} observes both $q$ and the conversation history $h$, autonomously generating a high-quality query description $q^*$, This aims to resolve ambiguities in user queries and reduce retrieval bias caused by incomplete or inaccurate expressions:
\begin{equation}
q^* = \text{QR}(q, h)
\end{equation}
\begin{equation}
\mathcal{T}_{q^*} = R_C(q^*, 3)
\end{equation}
where $\text{QR}(\cdot,\cdot)$ denotes the \textit{Query Refinement Agent}.

\paragraph{2) Divergent Thinking}
The \textit{Divergent Thinking Agent} then analyzes the initially retrieved knowledge snippets $\mathcal{T}_{q^*}$, generating $m$ divergent queries $\mathcal{Q}_{\text{DT}}$ from different perspectives related to $q^*$. These expanded queries broaden the problem space across various dimensions while remaining relevant to the original intent:
\begin{equation}
\mathcal{Q}_{\text{DT}} = \{q_1, q_2, \dots, q_m\} = \text{DT}(q^*, \mathcal{T}_q^*)
\end{equation}
\begin{equation}
\mathcal{T}_{\mathcal{Q}_{\text{DT}}} = \bigcup_{i=1}^{m} R_C(q_i, 3)
\end{equation}
where $\text{DT}(\cdot,\cdot)$ represents the \textit{Divergent Thinking Agent}, and $\mathcal{T}_{\mathcal{Q}_{\text{DT}}}$ denotes the deduplicated knowledge snippets retrieved from all divergent queries.

\paragraph{3) Knowledge Reasoning}
Finally, the \textit{Answer Generation Agent} synthesizes all retrieved knowledge snippets to address the original query:
\begin{equation}
a = \text{AG}(q, \mathcal{T}_q^* \cup \mathcal{T}_{\mathcal{Q}_{\text{DT}}})
\end{equation}
where $a$ represents the final answer, and \text{AG} denotes the \textit{Answer Generation Agent} guided by chain-of-thought reasoning prompts\cite{wei2022chain}.

In KidneyTalk-open's implementation, we perform additional filtering on $\mathcal{T}_q^* \cup \mathcal{T}_{\mathcal{Q}_{\text{DT}}}$: 1) Threshold filtering removes knowledge snippets with distance scores $s > 0.5$; 2) An agent evaluates whether each snippet genuinely contributes to answering the query (providing rationales for user reference), retaining only approved snippets.

\section{Usage Instructions and Examples}
\label{sec:usage}

This section introduces the hardware requirements and installation process of KidneyTalk-open, as well as the construction of a knowledge base based on multiple CKD (Chronic Kidney Disease) diagnosis and treatment guideline documents. It then demonstrates KidneyTalk-open's document-based question-answering functionality through a medical query.

\subsection{Hardware Requirements and Installation}

KidneyTalk-open uses Mac Mini (M4) as the reference device - the most affordable model in Apple's desktop lineup featuring a 10-core CPU, 10-core GPU, 16GB unified memory, and 256GB SSD. The unified memory architecture of Apple M-series chips enables cost-effective AI model inference, achieving approximately 20 tokens/second inference speed for 7B parameter models. KidneyTalk-open defaults to using DeepSeek-R1:7b and BGE-M3 models.

The KidneyTalk-open installation package is freely available on our website\footnote{\url{https://github.com/PKUDigitalHealth/KidneyTalk-open/releases}}. KidneyTalk-open relies on Ollama as the local model inference engine, requiring separate download from Ollama's official website\footnote{\url{https://ollama.com/download}}. Once Ollama is installed, it can run automatically in the background without further operation. Launching KidneyTalk-open reveals the model download interface where users can click \textit{Download Models} to automatically all default models.

\subsection{Medical Knowledge Database Construction from Documentations}

\begin{figure}[htbp]
    \centering
    \includegraphics[width=0.65\textwidth]{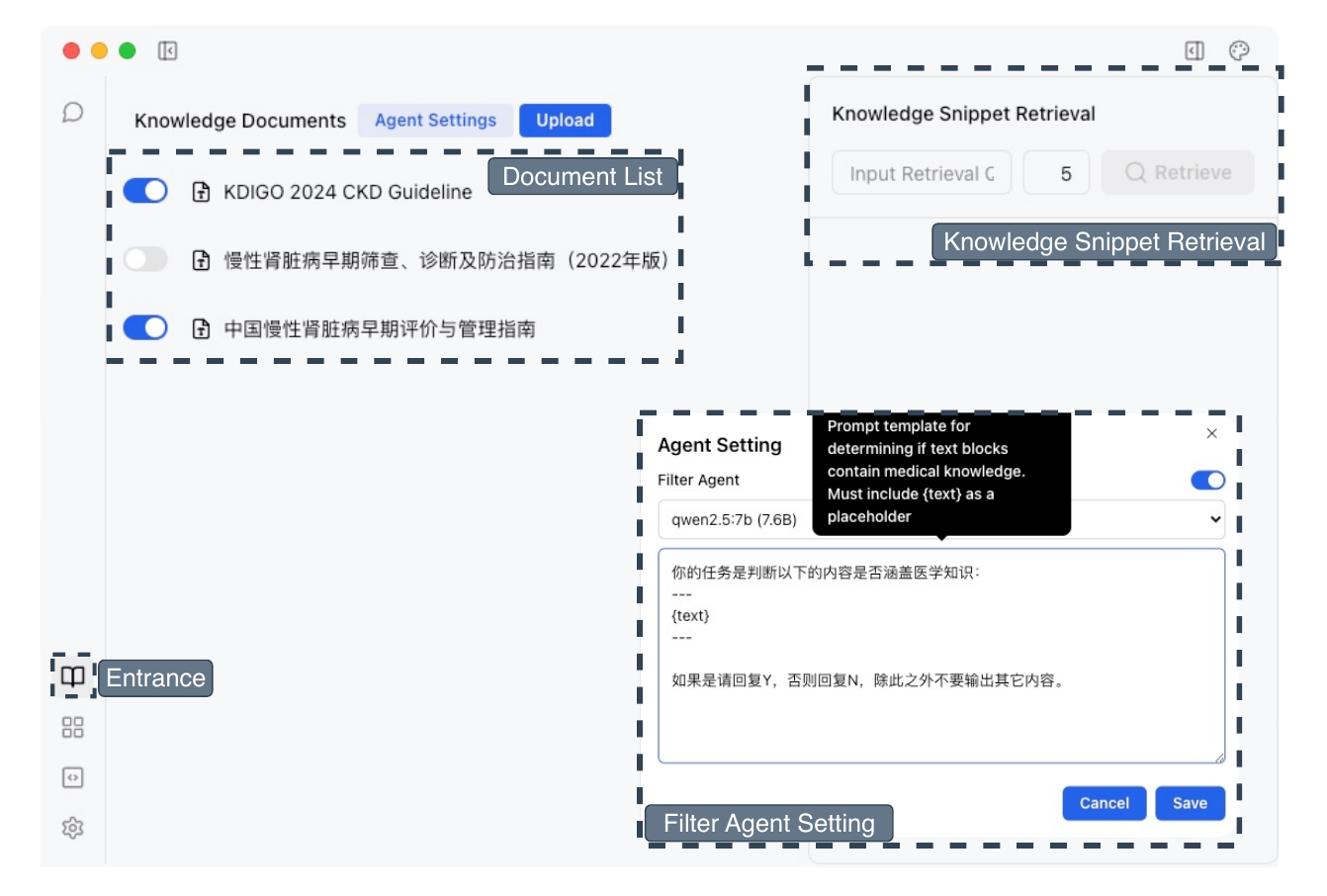}
    \caption{\textbf{Overview of KB construction page}. \textbf{Entrance}: This page is accessed by clicking the book icon. \textbf{Filter Agent Setting}: Although the construction is fully automated, users are allowed to customize the prompt of the \textit{Filter} in agent settings panel. This is used to intelligently control which knowledge snippets in documents will be embedded into the KB. \textbf{Document List}: Users are required to upload documents by clicking the \textit{Upload} button for building the KB. The documents existing in the KB will be listed in this area. If the document is "on" state, indicating the document will be retrieved to enhance the LLM. \textbf{Knowledge Snippet Retrieval}: A convenient feature allowing users to quickly retrieve knowledge snippets from the KB. For more detailed about this feature, please refer to Figure \ref{fig:snippets}.}
    \label{fig:knowledge}
\end{figure}

As shown in Figure \ref{fig:knowledge}, we imported three CKD guidelines into KidneyTalk-open's knowledge database module. Two of these guidelines \cite{egokcqcci2022guidelines,for2023guidelines} are in Chinese, and one \cite{stevens2024kdigo} is in English.

Users can upload arbitrary PDF, TXT, or Markdown documents to the knowledge database. In addition, in order to automatically filter out useless information (such as literature citations or tables of contents) according to users' wishes, users can enable the \textit{Filter Agent} before uploading documents by clicking the \textit{Agent Settings} button. For example, the default prompt of this agent is: \textit{Your task is to determine whether the following content contains medical knowledge...}. The technical details are described in section \ref{sec:filter_agent}.

\begin{figure}[htbp]
    \centering
    \includegraphics[width=0.85\textwidth]{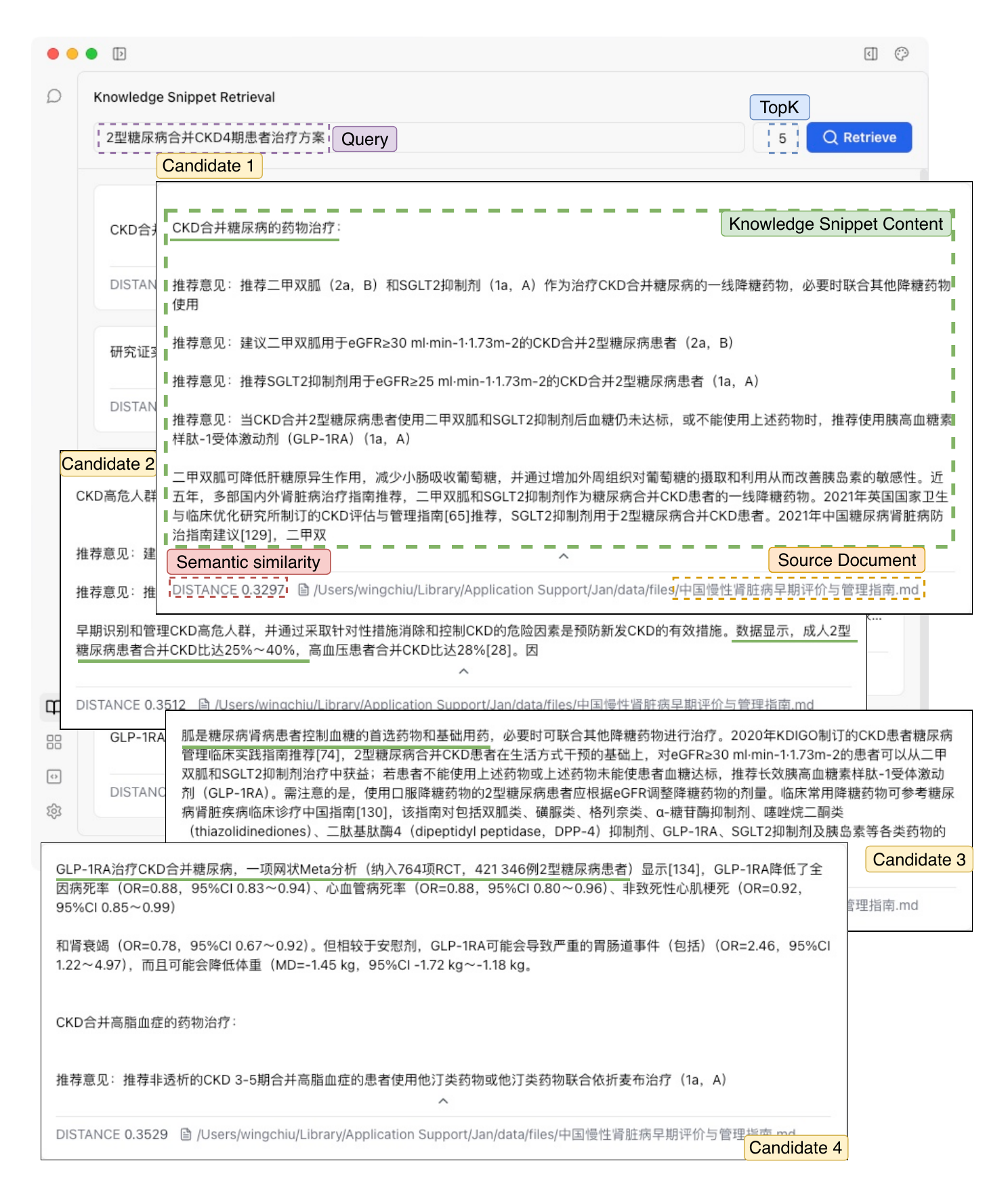}
    \caption{\textbf{Usage Example of the Knowledge Snippet Retrieval}. \textbf{Query}: We use the query \textit{"Treatment plans for patients with type 2 diabetes mellitus combined with stage 4 CKD"} as an example to demonstrate this function. \textbf{TopK}: We expects the KB to return the specified number knowledge snippets that are the most semantically similar to the query. \textbf{Candidate Knowledge Snippets}: We refer to the knowledge snippets returned by the retrieval as candidates. For visual ease, we display these candidates in the form of floating cards. Each candidate consists of content, semantic similarity, and source document. \textbf{Knowledge Snippet Content}: We use green underlines to mark the sentences related to the query, including descriptions of various treatment drugs for DKD and lifestyle and dietary patterns. \textbf{Semantic Similarity}: Distance represents the semantic similarity degree between the query and the content. It is calculated using cosine similarity formula. The smaller the value, the more similar they are. \textbf{Source Document}: The document path where the candidate is sourced from.}
    \label{fig:snippets}
\end{figure}

\subsection{Knowledge Snippets Retrieval from Medical Documentations}
KidneyTalk-open provides the \textit{Knowledge Snippet Retrieval} feature, allowing users to conduct semantic retrieval directly based on the knowledge database. As shown in Figure \ref{fig:snippets}, when querying \textit{Treatment plans for patients with type 2 diabetes and CKD4}, multiple knowledge snippets related to this query were returned. These snippets encompassed drug selections, such as the recommended use of SGLT2 inhibitors and GLP-1RA drugs. It should be noted that these snippets were sourced from the previously imported CKD guidelines. Moreover, clear traceability paths were provided along with the snippets, enabling users to conveniently verify the accuracy of the information. Each result was accompanied by a semantic similarity score, which helped users intuitively assess the relevance of the content. Snippets are sorted in ascending cosine distance order, with smaller values indicating approximately semantic proximity to the query. Thanks to this feature, users could directly perform semantic retrieval on the knowledge database constructed by KidneyTalk-open within seconds. This provided users with a precise, transparent, and rapidly traceable knowledge retrieval experience.

\subsection{Medical Q\&A with Knowledge Snippet References}

\begin{figure}[htbp]
    \centering
    \includegraphics[width=1.0\textwidth]{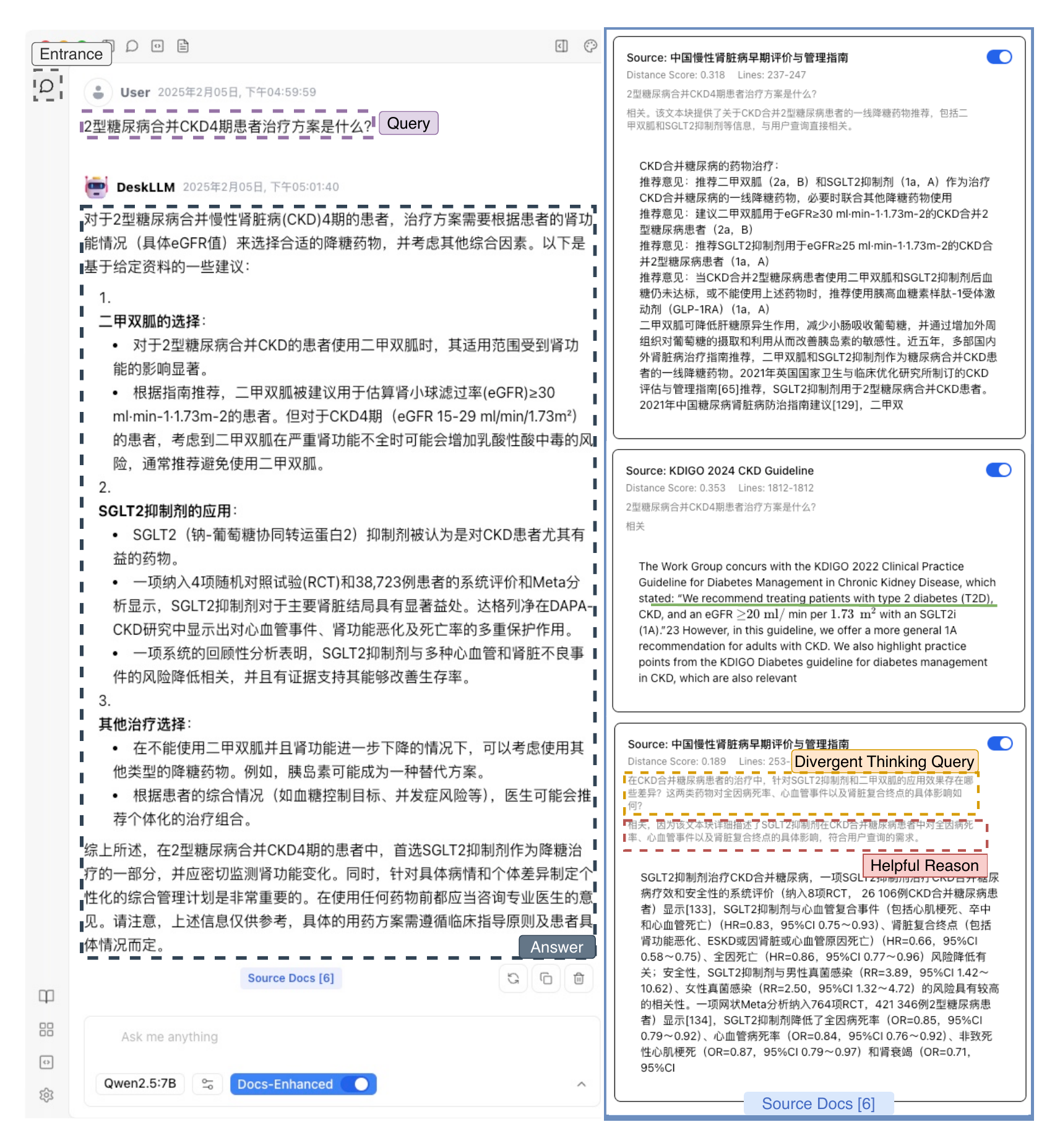}
    \caption{\textbf{Usage Example of the Chat Module}. \textbf{Entrance}: This page is accessed by clicking the \textit{chat} icon. \textbf{Query}: We use the query \textit{"What are the treatment plans for patients with type 2 diabetes and CKD4?"} as an example to demonstrate this function. \textbf{Answer}: The final answer of the model Qwen2.5:7B to the query when the knowledge database retrieval enhancement is enabled (enabled at the \textit{Docs-Enhanced}). \textbf{Source Docs}: KidneyTalk-open has successfully retrieved some helpful snippets from the knowledge database. \textbf{Divergent Thinking Query \& Helpful Reason}: Users can not only view the content of the snippets but also the process results of KidneyTalk-open, including the queries from the \textit{Divergent Thinking Agent} and the generated helpful reasons. The technical details are described in section \ref{sec:addrep}.}
    \label{fig:qwen}
\end{figure}

We integrated the agent technology of LLM (details in section \ref{sec:addrep}) into the Chat module of KidneyTalk-open to answer users' queries in a more natural way based on the knowledge database. The snippets returned by knowledge snippets retrieval are called candidate knowledge snippets. They are merely considered to be semantically similar to the query in the semantic space by the text embedding model, which doesn't necessarily mean they are actually helpful in answering the query. The Chat module, however, comprehensively processes, summarizes, and presents the retrieved snippets in a structured and readable manner through LLM, enabling users to obtain key information more efficiently. 

As shown in Figure \ref{fig:qwen}, we use the query - \textit{What are the treatment plans for patients with type 2 diabetes and CKD4?} for demonstration. For the query, the KidneyTalk-open not only extracted relevant snippets from multiple authoritative sources (such as the Chinese Chronic Kidney Disease Guidelines, the KDIGO Guidelines, etc.), but also summarized the applicable conditions, risk assessments, and evidence supports for different drugs (such as metformin, SGLT2 inhibitors, etc.). Compared with the scattered texts returned by knowledge snippets retrieval, the Chat module provided a more logical overview of the treatment plan, helping users quickly understand the core treatment strategies.

\begin{figure}[htbp]
    \centering
    \includegraphics[width=1.0\textwidth]{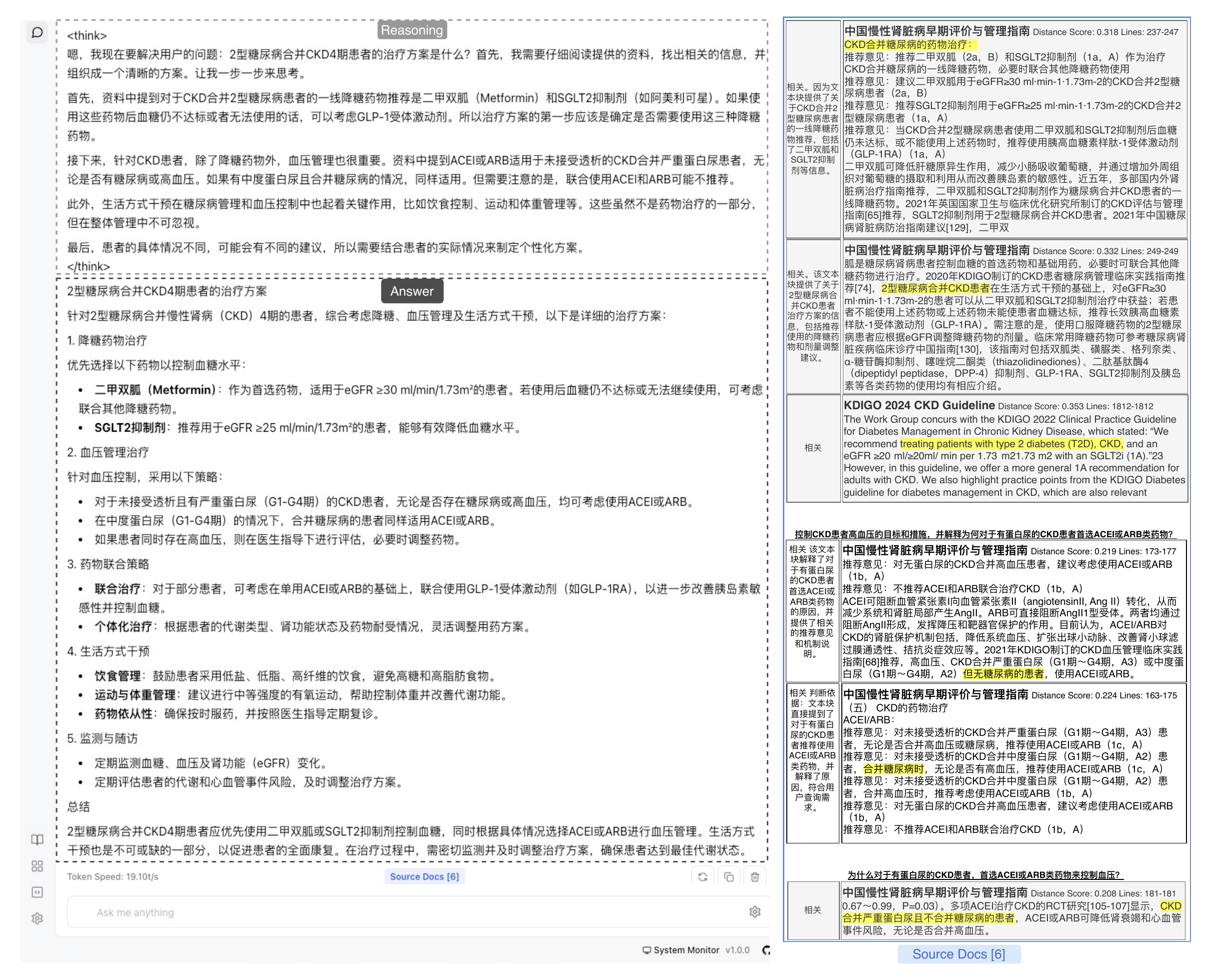}
    \caption{\textbf{Example Demonstrating the Reasoning Model DeepSeek-R1:7b .} We still use the same query as in figure \ref{fig:qwen} as an example to validate the reasoning model DeepSeek-R1 on KidneyTalk-open.The superiority of the reasoning model is demonstrated by comparing it with general model Qwen2.5:7b (as shown in Figure \ref{fig:qwen}). \textbf{Reasoning}: The content wrapped in the \textless think\textgreater\textless/think\textgreater tags is the reasoning process of DeepSeek-R1:7b based on the snippets retrieved by the KidneyTalk-open. \textbf{Answer}: Compared with the answer given by Qwen2.5:7B, the answer provided by DeepSeek-R1:7b after reasoning shows that DeepSeek-R1:7b has a higher utilization rate of snippets. We recommend using the reasoning model as the base model, which may lead to more reliable answers. \textbf{Source Docs:} For convenience, we reformatted the 6 retrieved knowledge snippets and displayed them on the right of the picture, highlighting the texts related to the query.} 
    \label{fig:deepseek}
\end{figure}

We show the superiority of the reasoning model deployed on the desktop side for enhancing knowledge database question answering. Figure \ref{fig:deepseek} shows that with the enhancement of the reasoning model DeepSeek-R1:7b, based on the same query as those in Figure \ref{fig:qwen}, the system gave a more comprehensive reply to the treatment of patients with type 2 diabetes mellitus combined with CKD4, including not only hypoglycemic drug treatment plan but also blood pressure control strategy, drug combination therapy strategy, and lifestyle interventions, fully utilizing the retrieved snippets to organize the final answer. Meanwhile, before giving the final answer, DeepSeek-R1:7b explicitly displays its reasoning process (the part wrapped by the tags \textless think\textgreater\ and \textless/think\textgreater), significantly increasing the transparency of LLM reasoning, which is particularly important in medical scenarios.

Compared with knowledge snippets retrieval, the advantages of the Chat module are as follows: 1) It can automatically screen and integrate multiple relevant knowledge snippets, relieving users from the burden of splicing and understanding fragmented information by themselves; 2) It can present answers in a logical and structured manner, making the information more intuitive and understandable; 3) It can interpret queries more accurately in combination with the context, avoiding directly returning redundant or incompletely matching content; 4) The traceability function is enhanced. Users can not only expand to view the content of the original documents on which the answers are based but also refer to the reasons why the agent considers the content helpful for answering, ensuring the traceability, reliability, and readability of the information. Therefore, the Chat module of KidneyTalk-open, which is based on medical document question answering, can effectively improve users' retrieval efficiency and the value of information utilization in scenarios such as medical knowledge answering and clinical decision support.

\section{Validation and Comparison}
\label{sec:val}

\subsection{Validation of AddRep}

The purpose of KidneyTalk-open is to provide users with LLM-based Q\&A capabilities enhanced by medical documents. Consistent with other medical RAG research \cite{singhal2023large,luo2024development,wu2024medical}, we have evaluated the effectiveness and robustness of KidneyTalk-open's core algorithm, AddRep, using medical exam multiple-choice questions (MCQs). These questions are not only a measure of the RAG system's ability to retrieve medical knowledge but also a test of the base model's medical reasoning capabilities. For this evaluation, we selected a challenging set of Chinese nephrology medical exam MCQs (CNME-MCQ) to validate AddRep. The CNME-MCQ dataset comprises 1,455 nephrology-related multiple-choice questions designed to assess the professional competence of medical students. The questions cover 31 different kidney diseases, with the top three in terms of question volume being glomerular diseases (328 questions), urinary tract infections, renal tubulointerstitial diseases, renal vascular diseases, and cystic kidney diseases (190 questions), and acute kidney failure (127 questions). Additionally, the questions are categorized into four types based on difficulty, from easiest to most difficult: \textit{A1/A2}, \textit{A3/A4}, \textit{X}, and \textit{Case Study}. The distribution and descriptions of each question type are provided in Table \ref{tab:nmemcq}. Regarding the retrievable corpus, in addition to the three CKD diagnosis and treatment guidelines, we incorporated four authoritative Chinese nephrology textbooks, ultimately constructing a knowledge database containing over 20,000 knowledge snippets.

\begin{table}[b]
    \caption{\textbf{Question Types in CNME-MCQ.} \textbf{A1/A2}: Short Single-CQs testing basic knowledge recall/understanding. \textbf{A3/A4}: Interrelated clinical Single-CQs assessing case analysis/application. \textbf{X}: Multi-CQs evaluating comprehensive knowledge/detail discrimination. \textbf{Case Study}: Complex scenarios with 51 Single-CQs + 170 Multi-CQs testing clinical reasoning/decision-making through logical progression. The total number is 1,455, and the difficulty of the types is gradually increasing.}   
    \centering
    \begin{tabular}{cccc}
            \hline
            A1/A2 & A3/A4 & X & Case Study\\  
            \hline
             714 & 314 & 206 & 221 \\ 
            \hline
            \end{tabular}
    \label{tab:nmemcq}
\end{table}

We selected DeepSeek-R1:7b as the base model for AddRep and designed two Baselines for comparison with AddRep: 1) \textit{Baseline}, which involves directly querying DeepSeek-r1:7b; 2) \textit{Baseline+RS}, which directly enhances the base model with knowledge snippets retrieved from the HNSW vector library. To more closely align with real-world medical information needs, we adopted the medical RAG system evaluation setup proposed by MIRAGE \cite{xiong2024benchmarking}. In this setup, when retrieving knowledge snippets, only the question stem is used as the initial query to increase the difficulty of retrieval. Additionally, to assess the robustness of above methods, we will calculate the model's rejection rate \cite{yu2024evaluation}. By prompting the model to refrain from answering uncertain multiple-choice questions, we can suppress the model's "hallucination" issue.

\begin{figure}[htbp]
    \centering
    \includegraphics[width=0.8\textwidth]{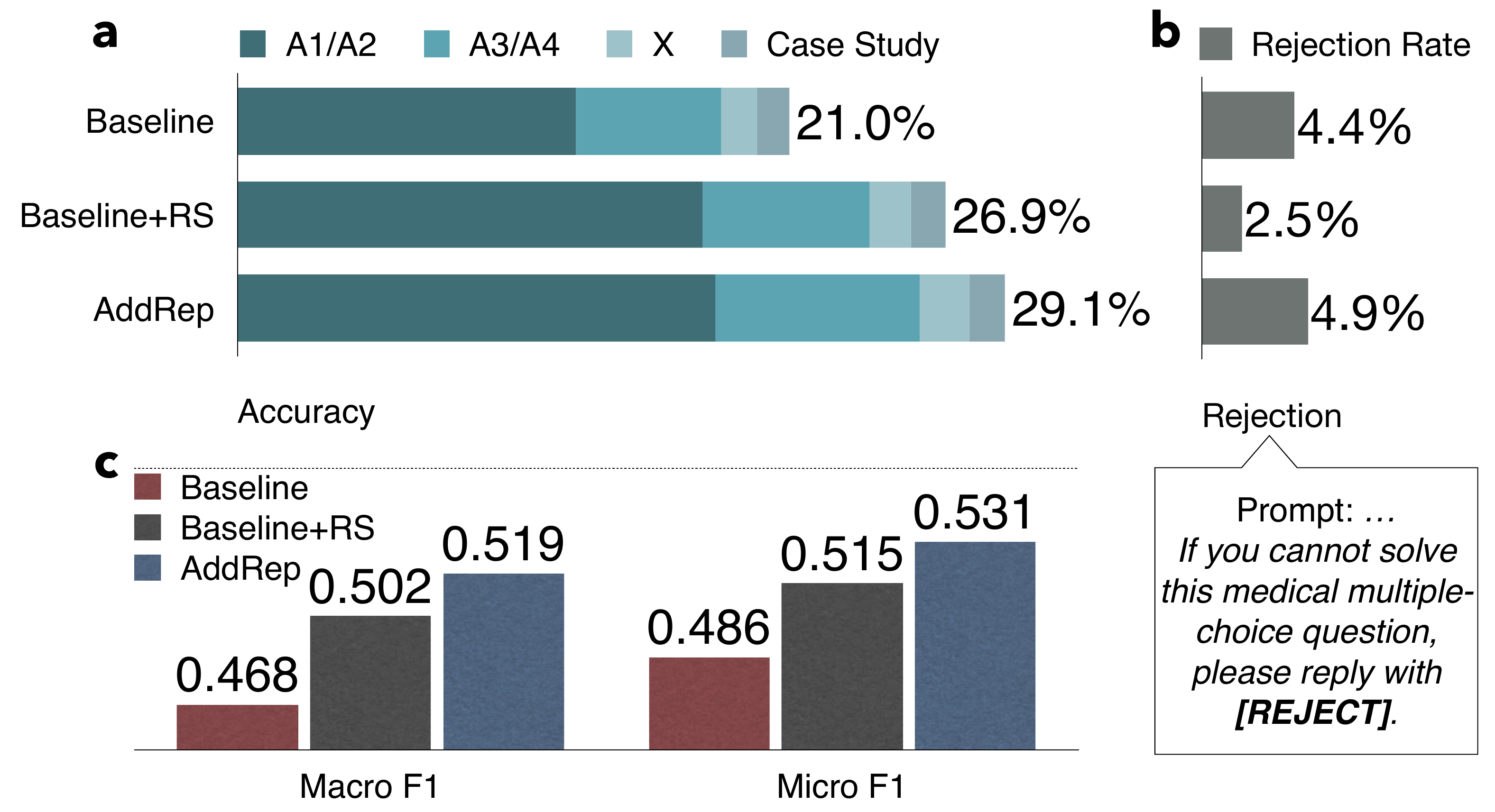}
    \caption{\textbf{Evaluation of AddRep on CNME-MCQ}. We use DeepSeek-R1:7b as the base model to verify the effectiveness of our proposed method AddRep. \textit{Baseline} means directly querying the base model, \textit{Baseline+RS(Retrieved Snippets)} is enhancing it with retrieved knowledge snippets from HNSW. \textbf{a} shows the overall accuracies of the three methods. \textit{A1/A2}, \textit{A3/A4}, \textit{X}, and \textit{Case Study} are question types of increasing difficulty. \textbf{b}: Our designed prompt asks the base model to judge if it can answer the question. If not, it replies \textit{[REJECT]}, suppressing hallucinations. AddRep has the highest accuracy(29.1\%) despite the highest rejection rate(4.9\%), indicating strong robustness. \textbf{c} presents the F1-scores. Micro F1 considers the ratio of single- and multiple-choice questions, unlike Macro F1.}
    \label{fig:addrep}
\end{figure}

As shown in Figure \ref{fig:addrep}, the experimental results demonstrate the superior performance of AddRep compared to the baseline methods. Specifically, AddRep achieved an overall accuracy of 29.1\%, significantly outperforming both the \textit{Baseline} (21.0\%) and \textit{Baseline+RS} (26.9\%) methods across all question types. This indicates that AddRep effectively enhances the base model's ability to retrieve and utilize relevant medical knowledge, thereby improving its reasoning capabilities in answering complex nephrology questions. In terms of rejection rate, AddRep exhibited a higher rejection rate of 4.9\% compared to the \textit{Baseline} (4.4\%) and \textit{Baseline+RS} (2.5\%). This higher rejection rate reflects AddRep's robustness in avoiding hallucinations by refraining from answering questions when the evidence is insufficient, thus ensuring higher reliability in its responses. Furthermore, the F1-scores, which are a harmonic mean of precision and recall, highlight AddRep's consistent improvement over the baseline methods. The Macro F1 score for AddRep was 0.519, while the Micro F1 score was 0.531, both of which were higher than those of the \textit{Baseline} and \textit{Baseline+RS}. This suggests that AddRep not only performs well on individual question types but also maintains a strong overall performance across the entire dataset, effectively balancing the ratio of single- and multiple-choice questions.

In summary, the experimental results clearly demonstrate that AddRep significantly enhances the base model's performance on the CNME-MCQ dataset, showcasing its effectiveness and robustness in a challenging medical examination context.

\subsection{Comparison with Mainstream LLMs Localized Applications}

\begin{figure}[htbp]
    \centering
    \includegraphics[width=1.0\textwidth]{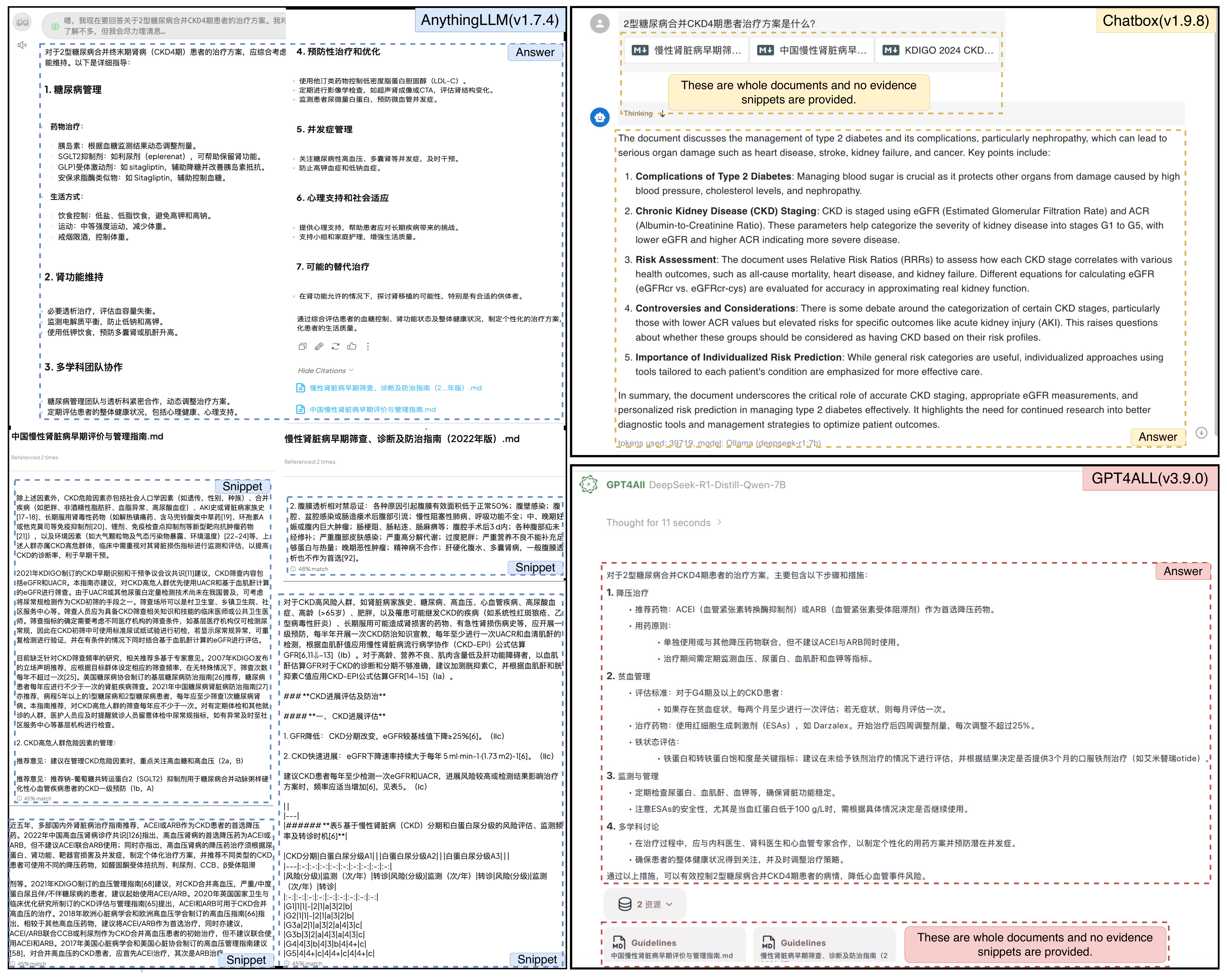}
    \caption{\textbf{Comparative Analysis of LLM localized Applications}. This study conducts a systematic evaluation of KidneyTalk-open (performance illustrated in Fig.\ref{fig:deepseek}) versus three mainstream applications on guideline-enhanced nephrology Q\&A tasks. A standardized control protocol was implemented: uniform deployment of DeepSeek-R1:7b base model, construction of knowledge bases from authoritative nephrology guidelines, and identical medical query ("What is the treatment plan for patients with type 2 diabetes mellitus combined with stage 4 CKD?"). \textbf{AnythingLLM}, while featuring comparable knowledge provenance feature to KidneyTalk-open, demonstrated inadequate retrieval of evidence related to Diabetic Kidney Disease (DKD) therapeutic protocols, resulting in clinically unsupported responses. \textbf{Chatbox}, constrained by inherent architectural limitations preventing local knowledge database construction, exhibited content truncation when attempting full-context integration of three guidelines, ultimately retaining only the terminal English guideline snippet. This caused non-native linguistic output with generalized content, indicating failure in DKD-specific information retrieval. \textbf{GPT4All}, despite supporting local knowledge database development, lacked knowledge provenance mechanisms. Its retrieval process omitted critical hypoglycemic medication protocols, generating recommendations deficient in essential clinical components.}
    \label{fig:others}
\end{figure}

We compared KidneyTalk-open with three mainstream LLMs localized applications on guideline-enhanced nephrology Q\&A tasks, including AnythingLLM\footnote{\url{https://anythingllm.com/}}, Chatbox\footnote{\url{https://chatboxai.app/en#}} and GPT4ALL\cite{gpt4all}. All three applications are full-stack application that supports private deployment and can transform documents into context, having functions relatively close to those of KidneyTalk-open. To ensure fairness, we still imported three CKD guidelines\cite{egokcqcci2022guidelines,for2023guidelines,stevens2024kdigo} into these applications as knowledge database or context. We used the same base model (DeepSeek-R1:7b) and then used the same medical query: "What is the treatment plan for patients with type 2 diabetes mellitus combined with stage 4 CKD?" to verify them. We compared the document retrieval capabilities, information integration abilities and answer quality of KidneyTalk-open and these applications. The results of KidneyTalk-open are shown in Figure \ref{fig:deepseek}, and the results of other applications are shown in Figure \ref{fig:others}.

To comprehensively evaluate the performance of various applications in generating DKD treatment plans, we established six essential clinical dimensions that should be addressed based on guideline recommendations: glycemic control medication guidance, blood pressure management strategies, lipid regulation protocols, lifestyle interventions, monitoring recommendations, and personalization based on CKD staging. These criteria were systematically analyzed through both document retrieval verification and final answer evaluation, as detailed in Table \ref{tab:comparison}. \textbf{1) Glycemic Control Guidance:} AnythingLLM failed to retrieve relevant information, resulting in generic suggestions about insulin and SGLT2 inhibitors without specific eGFR-adaptive recommendations. Both Chatbox and GPT4All exhibited unknown retrieval status and provided no concrete guidance. KidneyTalk-open successfully extracted guideline-based protocols, specifying metformin use ($\text{eGFR} \geq 30 \, \text{ml/min/1.73m}^2$) and SGLT2 inhibitors ($\text{eGFR} \geq 25 \, \text{ml/min/1.73m}^2$) as first-line therapies. \textbf{2) Blood Pressure Management:} While AnythingLLM retrieved relevant data, it failed to generate coherent responses, potentially due to useless information overload causing critical content loss \cite{liu2024lost}. Chatbox showed undefined retrieval status with no substantive answers. GPT4All unexpectedly provided accurate ACEI/ARB recommendations despite unclear retrieval processes. KidneyTalk-open demonstrated superior performance by offering staged management strategies aligned with CKD progression. \textbf{3) Lipid Control Protocols:} None of the applications successfully addressed this dimension, reflecting inherent challenges in cross-domain retrieval. Although lipid management is crucial for CKD patients, the absence of explicit diabetes mentions in guideline sections likely hindered effective knowledge extraction, requiring advanced medical reasoning beyond current capabilities. \textbf{4) Lifestyle Interventions:} Notably, AnythingLLM generated appropriate lifestyle suggestions despite lacking retrieval evidence, indicating potential base model memorization. KidneyTalk-open provided guideline-driven recommendations for low-sodium diets and moderate exercise, while Chatbox and GPT4All failed to address this aspect. \textbf{5) Monitoring Recommendations:} A paradoxical pattern emerged where AnythingLLM retrieved but omitted monitoring details, while KidneyTalk-open conversely synthesized monitoring advice from retrieved blood pressure/glucose information. GPT4All generated monitoring suggestions without clear retrieval basis. \textbf{6) Treatment Personalization:} Only KidneyTalk-open achieved true personalization by implementing CKD stage-specific medication adjustments. The three comparison applications provided generic advice regardless of disease progression, underscoring the critical importance of precise knowledge retrieval coupled with contextual reasoning. This multi-dimensional analysis reveals fundamental differences in architecture effectiveness. While basic retrieval functionality exists across platforms, KidneyTalk-open's integrated retrieval-reasoning pipeline demonstrates superior clinical utility through: 1) Precise knowledge anchoring with provenance tracking, 2) Context-aware information synthesis, and 3) Staged therapeutic recommendations adhering to nephrology practice guidelines.

\begin{table}[t]
\caption{\textbf{A comparative analysis of the performance of the responses provided by the four applications when optimizing the diabetic kidney disease (DKD) treatment plan based on guidelines-enhanced.} The table analyzes the performance of AnythingLLM, Chatbox, GPT4All(as shown in figure\ref{fig:others}), and KidneyTalk-open (as shown in \ref{fig:deepseek}) from six dimensions, including guidance for blood glucose control, guidance for blood pressure control, guidance for blood lipid control, lifestyle tips, monitoring tips and personalization (personalization based on CKD staging). \textbf{Retrieved} indicates whether the relevant knowledge is returned by the retrieval component, \textbf{Answered} indicates whether the final answer covers the content of this dimension (\textcolor{mygre}{Yes} represents yes, \textcolor{myred}{No} represents no, and \textcolor{mygra}{Unknown} represents unknown).}
\label{tab:comparison}
\centering
\small
\setlength{\tabcolsep}{4pt}
\begin{tabularx}{\textwidth}{@{}l *{8}{>{\centering\arraybackslash}X} @{}}
\toprule
 & \multicolumn{2}{c}{\textbf{AnythingLLM}} & \multicolumn{2}{c}{\textbf{Chatbox}} & \multicolumn{2}{c}{\textbf{GPT4All}} & \multicolumn{2}{c}{\textbf{KidneyTalk-open}} \\
\cmidrule(lr){2-3} \cmidrule(lr){4-5} \cmidrule(lr){6-7} \cmidrule(lr){8-9}
\textbf{Dimensions} & Retrieved & Answered & Retrieved & Answered& Retrieved & Answered& Retrieved & Answered\\
\midrule
Guidance for BG   & \textcolor{myred}{No}  & \textcolor{myred}{No}  & \textcolor{mygra}{Unknow} & \textcolor{myred}{No} & \textcolor{mygra}{Unkonw} & \textcolor{myred}{No}  & \textcolor{mygre}{Yes} & \textcolor{mygre}{Yes} \\
Guidance for BP   & \textcolor{myred}{No}  & \textcolor{myred}{No}  & \textcolor{mygra}{Unknow} & \textcolor{myred}{No}  & \textcolor{mygra}{Unkonw} & \textcolor{mygre}{Yes} & \textcolor{mygre}{Yes} & \textcolor{mygre}{Yes} \\
Guidance for BL   & \textcolor{myred}{No}  & \textcolor{myred}{No}  & \textcolor{mygra}{Unknow} & \textcolor{myred}{No}  & \textcolor{mygra}{Unkonw} & \textcolor{myred}{No}  & \textcolor{myred}{No}  & \textcolor{myred}{No}  \\
Lifestyle Tips    & \textcolor{myred}{No}  & \textcolor{mygre}{Yes} & \textcolor{mygra}{Unknow} & \textcolor{myred}{No}  & \textcolor{mygra}{Unkonw} & \textcolor{myred}{No}  & \textcolor{mygre}{Yes} & \textcolor{mygre}{Yes} \\
Monitoring        & \textcolor{mygre}{Yes} & \textcolor{myred}{No}  & \textcolor{mygra}{Unknow} & \textcolor{myred}{No}  & \textcolor{mygra}{Unkonw} & \textcolor{mygre}{Yes} & \textcolor{myred}{No}  & \textcolor{mygre}{Yes} \\
Personalization   & \textcolor{myred}{No}  & \textcolor{myred}{No}  & \textcolor{mygra}{Unknow} & \textcolor{myred}{No}  & \textcolor{mygra}{Unkonw} & \textcolor{myred}{No}  & \textcolor{mygre}{Yes} & \textcolor{mygre}{Yes} \\
\bottomrule
\end{tabularx}
\end{table}

\section{Conclusion and Future Work}

KidneyTalk-open, as the first no-code private large language model (LLM) system for clinical scenarios, has successfully achieved the local integration of medical knowledge management and intelligent reasoning. By innovatively combining the open-source LLM deployment framework, semantic knowledge database construction, and multi-agent retrieval enhancement technologies, the system effectively solves three critical challenges in medical LLMs applications: 1) the conflict between cloud-based model services and patient privacy protection; 2) the deficiency in timely medical knowledge updates within general LLMs; 3) The dilemma of the mismatch between complex computer technical processes and the skills of clinical users. Experimental results on Chinese nephrology medical examination multiple-choice questions, along with case study comparisons against three mainstream medical applications, demonstrate that the system exhibits superior knowledge localization capabilities and evidence-based reasoning quality in professional scenarios such as kidney disease diagnosis and treatment. This advancement provides a novel tool for promoting the inclusive development of intelligent healthcare.

Although KidneyTalk-open has made remarkable progress in the field of medical knowledge QA, there are still improvement directions worthy of in-depth exploration: 1) Enhancement of complex document parsing. Currently, the system has limited processing capabilities for documents containing complex tables, mathematical formulas, and scanned PDF files. In the future, it is planned to integrate a localized PDF parsing model, such as MinerU\cite{wang2024mineruopensourcesolutionprecise}; 2) Expansion of multi-modal medical understanding. It is proposed to introduce visual language models, build a pipeline for analyzing laboratory test sheet images, enabling the system to automatically extract the values and clinical significance of test indicators, and combine with an electrocardiogram waveform analysis module to achieve multi-modal data fusion reasoning. Through continuous optimization, KidneyTalk-open is expected to develop into an important intelligent decision-making partner in clinical work, promoting the practice of precise diagnosis and treatment based on evidence-based medicine. The open-source implementation scheme of this study provides a reusable technical framework and design paradigm for the development of subsequent medical AI systems.

\section*{Acknowledgments}
We thank the contributors of the Jan project\cite{janhq} for developing a fully offline desktop-based alternative to ChatGPT, emphasizing complete privacy and local execution.
KidneyTalk-open will continue to comply with the AGPL-3.0 license, ensuring adherence to open-source principles and fostering further innovation in the field.




\printbibliography

@article{jeong2024limited,
  title={The Limited Impact of Medical Adaptation of Large Language and Vision-Language Models},
  author={Jeong, Daniel P and Mani, Pranav and Garg, Saurabh and Lipton, Zachary C and Oberst, Michael},
  journal={arXiv preprint arXiv:2411.08870},
  year={2024}
}

@inproceedings{dou2024loramoe,
  title={LoRAMoE: Alleviating world knowledge forgetting in large language models via MoE-style plugin},
  author={Dou, Shihan and Zhou, Enyu and Liu, Yan and Gao, Songyang and Shen, Wei and Xiong, Limao and Zhou, Yuhao and Wang, Xiao and Xi, Zhiheng and Fan, Xiaoran and others},
  booktitle={Proceedings of the 62nd Annual Meeting of the Association for Computational Linguistics (Volume 1: Long Papers)},
  pages={1932--1945},
  year={2024}
}

@article{malkov2018efficient,
  title={Efficient and robust approximate nearest neighbor search using hierarchical navigable small world graphs},
  author={Malkov, Yu A and Yashunin, Dmitry A},
  journal={IEEE transactions on pattern analysis and machine intelligence},
  volume={42},
  number={4},
  pages={824--836},
  year={2018},
  publisher={IEEE}
}

@article{luo2024development,
  title={Development and evaluation of a retrieval-augmented large language model framework for ophthalmology},
  author={Luo, Ming-Jie and Pang, Jianyu and Bi, Shaowei and Lai, Yunxi and Zhao, Jiaman and Shang, Yuanrui and Cui, Tingxin and Yang, Yahan and Lin, Zhenzhe and Zhao, Lanqin and others},
  journal={JAMA ophthalmology},
  volume={142},
  number={9},
  pages={798--805},
  year={2024},
  publisher={American Medical Association}
}

@article{trivedi2022interleaving,
  title={Interleaving retrieval with chain-of-thought reasoning for knowledge-intensive multi-step questions},
  author={Trivedi, Harsh and Balasubramanian, Niranjan and Khot, Tushar and Sabharwal, Ashish},
  journal={arXiv preprint arXiv:2212.10509},
  year={2022}
}

@article{ram2023context,
  title={In-context retrieval-augmented language models},
  author={Ram, Ori and Levine, Yoav and Dalmedigos, Itay and Muhlgay, Dor and Shashua, Amnon and Leyton-Brown, Kevin and Shoham, Yoav},
  journal={Transactions of the Association for Computational Linguistics},
  volume={11},
  pages={1316--1331},
  year={2023},
  publisher={MIT Press One Broadway, 12th Floor, Cambridge, Massachusetts 02142, USA~…}
}

@inproceedings{asai2023retrieval,
  title={Retrieval-based language models and applications},
  author={Asai, Akari and Min, Sewon and Zhong, Zexuan and Chen, Danqi},
  booktitle={Proceedings of the 61st Annual Meeting of the Association for Computational Linguistics (Volume 6: Tutorial Abstracts)},
  pages={41--46},
  year={2023}
}

@article{yang2024qwen2,
  title={Qwen2. 5 Technical Report},
  author={Yang, An and Yang, Baosong and Zhang, Beichen and Hui, Binyuan and Zheng, Bo and Yu, Bowen and Li, Chengyuan and Liu, Dayiheng and Huang, Fei and Wei, Haoran and others},
  journal={arXiv preprint arXiv:2412.15115},
  year={2024}
}

@article{liu2024deepseek,
  title={DeepSeek-V3 Technical Report},
  author={Liu, Aixin and Feng, Bei and Xue, Bing and Wang, Bingxuan and Wu, Bochao and Lu, Chengda and Zhao, Chenggang and Deng, Chengqi and Zhang, Chenyu and Ruan, Chong and others},
  journal={arXiv preprint arXiv:2412.19437},
  year={2024}
}

@article{dubey2024llama,
  title={The llama 3 herd of models},
  author={Dubey, Abhimanyu and Jauhri, Abhinav and Pandey, Abhinav and Kadian, Abhishek and Al-Dahle, Ahmad and Letman, Aiesha and Mathur, Akhil and Schelten, Alan and Yang, Amy and Fan, Angela and others},
  journal={arXiv preprint arXiv:2407.21783},
  year={2024}
}

@misc{ollama,
  author = "{Ollama}",
  title = "{Ollama}",
  howpublished = "\url{https://github.com/ollama/ollama}",
}

@misc{janhq,
  author = "{janhq}",
  title = "{Jan}",
  howpublished = "\url{https://github.com/janhq/jan}",
}

@misc{cortex,
  author = "{janhq}",
  title = "{Cortex}",
  howpublished = "\url{https://github.com/janhq/cortex.cpp}",
}

@misc{hnswlib,
  author = "{yoshoku}",
  title = "{hnswlib-node}",
  howpublished = "\url{https://github.com/yoshoku/hnswlib-node}",
}

@article{vaswani2017attention,
  title={Attention is all you need},
  author={Vaswani, A},
  journal={Advances in Neural Information Processing Systems},
  year={2017}
}

@misc{wang2024mineruopensourcesolutionprecise,
      title={MinerU: An Open-Source Solution for Precise Document Content Extraction}, 
      author={Bin Wang and Chao Xu and Xiaomeng Zhao and Linke Ouyang and Fan Wu and Zhiyuan Zhao and Rui Xu and Kaiwen Liu and Yuan Qu and Fukai Shang and Bo Zhang and Liqun Wei and Zhihao Sui and Wei Li and Botian Shi and Yu Qiao and Dahua Lin and Conghui He},
      year={2024},
      eprint={2409.18839},
      archivePrefix={arXiv},
      primaryClass={cs.CV},
      url={https://arxiv.org/abs/2409.18839}, 
}

@article{guo2025deepseek,
  title={Deepseek-r1: Incentivizing reasoning capability in llms via reinforcement learning},
  author={Guo, Daya and Yang, Dejian and Zhang, Haowei and Song, Junxiao and Zhang, Ruoyu and Xu, Runxin and Zhu, Qihao and Ma, Shirong and Wang, Peiyi and Bi, Xiao and others},
  journal={arXiv preprint arXiv:2501.12948},
  year={2025}
}

@article{singhal2025toward,
  title={Toward expert-level medical question answering with large language models},
  author={Singhal, Karan and Tu, Tao and Gottweis, Juraj and Sayres, Rory and Wulczyn, Ellery and Amin, Mohamed and Hou, Le and Clark, Kevin and Pfohl, Stephen R and Cole-Lewis, Heather and others},
  journal={Nature Medicine},
  pages={1--8},
  year={2025},
  publisher={Nature Publishing Group US New York}
}

@article{liu2025generalist,
  title={A generalist medical language model for disease diagnosis assistance},
  author={Liu, Xiaohong and Liu, Hao and Yang, Guoxing and Jiang, Zeyu and Cui, Shuguang and Zhang, Zhaoze and Wang, Huan and Tao, Liyuan and Sun, Yongchang and Song, Zhu and others},
  journal={Nature Medicine},
  pages={1--11},
  year={2025},
  publisher={Nature Publishing Group US New York}
}

@article{wei2022chain,
  title={Chain-of-thought prompting elicits reasoning in large language models},
  author={Wei, Jason and Wang, Xuezhi and Schuurmans, Dale and Bosma, Maarten and Xia, Fei and Chi, Ed and Le, Quoc V and Zhou, Denny and others},
  journal={Advances in neural information processing systems},
  volume={35},
  pages={24824--24837},
  year={2022}
}

@misc{langchainjs,
  author       = "langchain-ai",
  title        = "langchainjs",
  howpublished = "\url{https://github.com/langchain-ai/langchainjs}",
  year         = "2023"
}

@article{xiong2024benchmarking,
  title={Benchmarking retrieval-augmented generation for medicine},
  author={Xiong, Guangzhi and Jin, Qiao and Lu, Zhiyong and Zhang, Aidong},
  journal={arXiv preprint arXiv:2402.13178},
  year={2024}
}

@article{singhal2023large,
  title={Large language models encode clinical knowledge},
  author={Singhal, Karan and Azizi, Shekoofeh and Tu, Tao and Mahdavi, S Sara and Wei, Jason and Chung, Hyung Won and Scales, Nathan and Tanwani, Ajay and Cole-Lewis, Heather and Pfohl, Stephen and others},
  journal={Nature},
  volume={620},
  number={7972},
  pages={172--180},
  year={2023},
  publisher={Nature Publishing Group}
}

@article{wu2024medical,
  title={Medical Graph RAG: Towards Safe Medical Large Language Model via Graph Retrieval-Augmented Generation},
  author={Wu, Junde and Zhu, Jiayuan and Qi, Yunli},
  journal={arXiv preprint arXiv:2408.04187},
  year={2024}
}

@inproceedings{yu2024evaluation,
  title={Evaluation of retrieval-augmented generation: A survey},
  author={Yu, Hao and Gan, Aoran and Zhang, Kai and Tong, Shiwei and Liu, Qi and Liu, Zhaofeng},
  booktitle={CCF Conference on Big Data},
  pages={102--120},
  year={2024},
  organization={Springer}
}

@article{chan2024rq,
  title={Rq-rag: Learning to refine queries for retrieval augmented generation},
  author={Chan, Chi-Min and Xu, Chunpu and Yuan, Ruibin and Luo, Hongyin and Xue, Wei and Guo, Yike and Fu, Jie},
  journal={arXiv preprint arXiv:2404.00610},
  year={2024}
}

@misc{gpt4all,
  author = {Yuvanesh Anand and Zach Nussbaum and Brandon Duderstadt and Benjamin Schmidt and Andriy Mulyar},
  title = {GPT4All: Training an Assistant-style Chatbot with Large Scale Data Distillation from GPT-3.5-Turbo},
  year = {2023},
  publisher = {GitHub},
  journal = {GitHub repository},
  howpublished = {\url{https://github.com/nomic-ai/gpt4all}},
}

@article{egokcqcci2022guidelines,
  title={Guidelines for early screening, diagnosis, prevention and treatment of chronic kidney disease (2022 edition)},
  author={EGoKCQCCi, S},
  journal={Chinese Journal of Nephrology},
  volume={38},
  number={5},
  pages={453--464},
  year={2022}
}

@article{for2023guidelines,
  title={Guidelines for the early evaluation and management of chronic kidney disease in China},
  author={for Kidney, Chinese Preventive Medicine Association and others},
  journal={Zhonghua nei ke za zhi},
  volume={62},
  number={8},
  pages={902--930},
  year={2023}
}

@article{stevens2024kdigo,
  title={KDIGO 2024 clinical practice guideline for the evaluation and management of chronic kidney disease},
  author={Stevens, Paul E and Ahmed, Sofia B and Carrero, Juan Jesus and Foster, Bethany and Francis, Anna and Hall, Rasheeda K and Herrington, Will G and Hill, Guy and Inker, Lesley A and Kazanc{\i}o{\u{g}}lu, R{\"u}meyza and others},
  journal={Kidney international},
  volume={105},
  number={4},
  pages={S117--S314},
  year={2024},
  publisher={Elsevier}
}

@article{liu2024lost,
  title={Lost in the middle: How language models use long contexts},
  author={Liu, Nelson F and Lin, Kevin and Hewitt, John and Paranjape, Ashwin and Bevilacqua, Michele and Petroni, Fabio and Liang, Percy},
  journal={Transactions of the Association for Computational Linguistics},
  volume={12},
  pages={157--173},
  year={2024},
  publisher={MIT Press One Broadway, 12th Floor, Cambridge, Massachusetts 02142, USA~…}
}

@inproceedings{fan2024survey,
  title={A survey on rag meeting llms: Towards retrieval-augmented large language models},
  author={Fan, Wenqi and Ding, Yujuan and Ning, Liangbo and Wang, Shijie and Li, Hengyun and Yin, Dawei and Chua, Tat-Seng and Li, Qing},
  booktitle={Proceedings of the 30th ACM SIGKDD Conference on Knowledge Discovery and Data Mining},
  pages={6491--6501},
  year={2024}
}

@article{kresevic2024optimization,
  title={Optimization of hepatological clinical guidelines interpretation by large language models: a retrieval augmented generation-based framework},
  author={Kresevic, Simone and Giuffr{\`e}, Mauro and Ajcevic, Milos and Accardo, Agostino and Croc{\`e}, Lory S and Shung, Dennis L},
  journal={NPJ digital medicine},
  volume={7},
  number={1},
  pages={102},
  year={2024},
  publisher={Nature Publishing Group UK London}
}

\end{document}